\documentclass[9pt]{extarticle}
\usepackage{spconf,amsmath,graphicx}
\usepackage{amssymb}
\usepackage{multirow}
\usepackage{booktabs}
\usepackage{ctable}
\usepackage{stix}
\usepackage{cite}
\usepackage{amsmath,amssymb}
\usepackage{url}

\makeatletter
\renewcommand{\@makefnmark}{\hbox{\@textsuperscript{\fnsymbol{footnote}}}}
\makeatother

\title{DA-Mamba: Dialogue-Aware Selective State-Space Model for Multimodal Engagement Estimation}

\name{Shenwei Kang$^{1}$,Xin Zhang$^1$,Wen Liu$^1$, Bin Li$^{2\textasteriskcentered}$, Yujie Liu$^{3}$,  Bo Gao$^{3\textasteriskcentered}$}
\vspace{-3em}
\address{$^1$Communication University of China, Beijing, China, \\$^2$Shenzhen Institute of Advanced Technology, Chinese Academy of Sciences‌,
\\ $^3$School of Information Engineering, Beijing Institute of Graphic Communication}

\begin{document}
%
\maketitle
\begin{abstract}
Human engagement estimation in conversational scenarios is essential for applications such as adaptive tutoring, remote healthcare assessment, and socially aware human--computer interaction. Engagement is a dynamic, multimodal signal conveyed by facial expressions, speech, gestures, and behavioral cues over time. In this work we introduce \textbf{DA-Mamba}, a dialogue-aware multimodal architecture that replaces attention-heavy dialogue encoders with Mamba-based selective state-space processing to achieve linear time and memory complexity while retaining expressive cross-modal reasoning. We design a Mamba dialogue-aware selective state-space model composed of three core modules: a Dialogue-Aware Encoder, and two Mamba-based fusion mechanisms: Modality-Group Fusion and Partner-Group Fusion, these modules achieve expressive dialogue understanding. Extensive experiments on three standard benchmarks (NoXi, NoXi-Add, and MPIIGI) show that DA-Mamba \textbf{surpasses prior state-of-the-art (SOTA) methods} in concordance correlation coefficient (CCC), while reducing training time and peak memory; these gains enable processing much longer sequences and facilitate real-time deployment in resource-constrained, multi-party conversational settings. The source code will be available at: \url{https://github.com/kksssssss-ssda/MMEA}.

\end{abstract}   
\vspace{-0.2cm}
\section{Introduction}
\label{sec:intro}
\footnotetext{$^{*}$Corresponding author.}
Human engagement estimation has emerged as a pivotal research area in affective computing and social signal processing, with broad applications spanning from educational technology and healthcare to human--computer interaction systems~\cite{cafaro2021noxi, li2024towards, Li2024_DAT}.
The ability to automatically assess an individual's level of attention, interest, and behavioral involvement during conversations enables the development of responsive and adaptive artificial systems \cite{muller2021multimediate,Muller2024_MultiMediate}. Engagement is inherently multimodal, expressed through a combination of facial expressions, vocal characteristics, gestures, and other non-verbal cues that evolve continuously over time~\cite{zadeh2017tensor,tan2019lxmert}.

Early computational approaches for engagement estimation predominantly relied on recurrent neural architectures.
Methods utilizing Recurrent Neural Networks (RNNs)~\cite{elman1990finding} and Long Short-Term Memory networks (LSTMs)~\cite{hochreiter1997long} processed sequential multimodal features to capture temporal dynamics in participant behavior.
While these models demonstrated promising results in short-term engagement tracking, they often struggled with capturing long-range dependencies and suffered from vanishing gradient problems, limiting their effectiveness in extended conversational contexts.

The introduction of Transformer-based models~\cite{vaswani2017attention} marked a significant advancement, leveraging self-attention mechanisms to model longer contextual relationships more effectively.
Transformers enabled better handling of multimodal feature interactions through dedicated encoding layers for different modalities~\cite{zadeh2017tensor,tan2019lxmert}. However, standard Transformers typically processed features through simple concatenation or early fusion strategies without fully exploiting the intrinsic structure within and across modalities~\cite{zadeh2017tensorEMNLP,li2022emotion}.

To address these limitations, the Dialogue-Aware Transformer (DAT) framework was proposed, representing a substantial step forward by explicitly modeling both cross-modal and cross-participant interactions. DAT introduced dedicated Modality-Group Fusion modules that independently process audio and visual features before integration, reducing redundancy and enhancing representation learning. Furthermore, its Dialogue-Aware Encoder employed cross-attention mechanisms to incorporate the behavioral cues of the conversational partner, significantly improving the accuracy of the engagement estimation by contextualizing the behavior of the target participant within dialogue dynamics~\cite{muller2021multimediate,Muller2024_MultiMediate,Li2024_DAT}.

\begin{figure*}[t!]
  \centering
  \includegraphics[width=.95\linewidth]{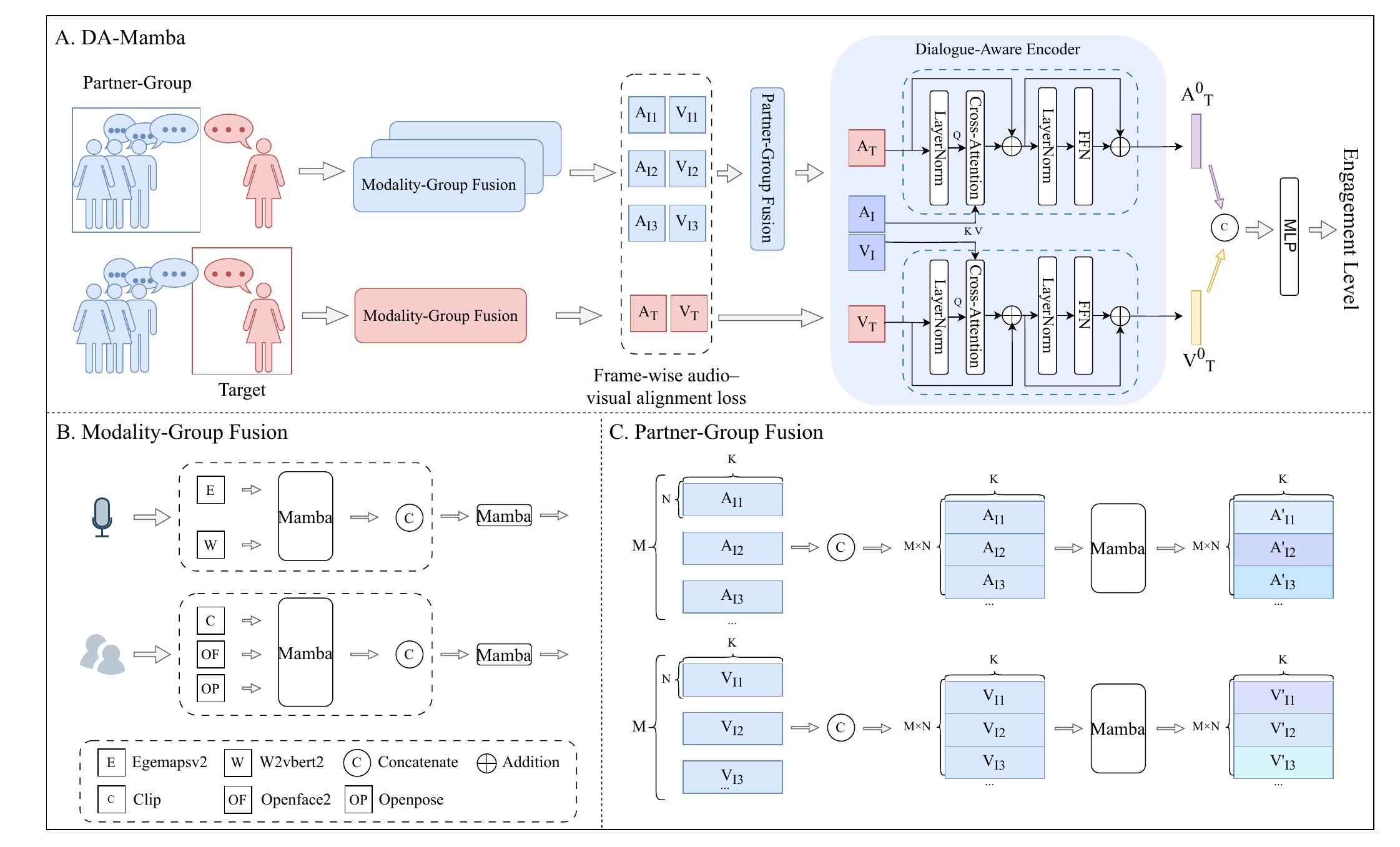}
    \vspace{-0.3cm}
  \caption{Overview of the DA-Mamba architecture. 
  (A) Main pipeline: per-person feature extraction $\rightarrow$ modality-group Mamba $\rightarrow$ target alignment $\rightarrow$ partner-context assembly $\rightarrow$ cross-attention $\rightarrow$ frame-score regression. 
  (B) Per-person modality grouping and Mamba processing: audio-group (Ege+W2v) and visual-group (CLIP+OF+OF2) are fed into two \emph{independent} Mamba stacks. 
  (C) Partner-context assembly: partners' outputs are concatenated along the frame axis and processed by a \emph{context Mamba}; the resulting context embeddings are then used in modality-wise cross-attention with the target frames.}
  \label{fig:overall_arch}
  \vspace{-0.3cm}
\end{figure*}

Despite these advancements, DAT and similar attention-based architectures face fundamental scalability challenges. The self-attention mechanism's computational and memory requirements grow quadratically with sequence length, making these models prohibitively expensive for long conversations and multi-party settings where processing extended temporal contexts is essential~\cite{vaswani2017attention}. This limitation becomes particularly problematic in real-world applications where conversations may last for hours or involve multiple participants, creating significant barriers to practical deployment~\cite{muller2021multimediate}.

To overcome these constraints while preserving the benefits of contextual modeling, we propose \textbf{DA-Mamba}, a dialogue-aware, multimodal architecture that integrates selective state-space models (SSMs) into the engagement estimation pipeline. Building on recent advances in selective SSM and Mamba-style sequence modelling~\cite{gu2021efficiently,gu2023mamba}, DA-Mamba replaces costly quadratic attention with efficient, linear-time SSM blocks while retaining the expressivity required for fine-grained cross-modal and cross-participant reasoning. Concretely, DA-Mamba combines modality-specific SSM stacks, a lightweight cross-modal initialization scheme, and a cross-modal Mamba fusion module to achieve both algorithmic scalability and strong representational power~\cite{zhu2024multimodalmamba,liu2024videomamba,wang2024visionmamba,liu2024multimodal,ma2024mamba,zhang2024vmamba,liu2024mamba,chen2024mm,rahmani2024mambamixer,wang2024multimodalmamba,li2024mamba}.

\textbf{Our main contributions:} a scalable selective-SSM backbone that hybridizes local chunked attention with selective SSM global branches for linear complexity and long-range modeling; modality-group and partner-group modeling via modality-specific SSM stacks and partner-context assembly for hierarchical fusion; and frame-wise cross-modal supervision using a symmetric InfoNCE loss and lightweight cross-attention for temporal alignment and robustness. Collectively, these enable practical, high-performance engagement estimation in resource-constrained real-time applications.
  \vspace{-0.1cm}
\section{The DA-Mamba Model}
\label{sec:DA-Mamba}
  \vspace{-0.1cm}
\subsection{Overview: DA-Mamba}
\label{sec:overview}
\vspace{-0.2cm}

The overall architecture of our proposed DA-Mamba framework is illustrated in Figure~\ref{fig:overall_arch}. The system processes multi-party conversations with $M$ participants to produce frame-level engagement predictions for a designated target participant $t$ through a structured pipeline: per-person feature extraction, modality-group Mamba processing, target alignment, partner-context assembly, cross-attention fusion, and frame-score regression. For each participant, frame-level features are grouped into audio-group (e.g., eGeMAPS, W2V-BERT) and visual-group (e.g., CLIP, OpenFace, OpenPose) representations, each processed by independent Mamba stacks to obtain deep modality-specific frame embeddings with frame-wise alignment loss enforcing cross-modal consistency. The embeddings from all $M-1$ conversational partners are concatenated temporally to form a long partner-context sequence, processed by a dedicated context Mamba module, and then fused with the target's embeddings via modality-wise cross-attention (audio-to-audio and visual-to-visual) to capture partner behavioral influences. Finally, a lightweight MLP head maps the enriched frame-aligned interaction features to the final engagement score $\hat{y}_t \in \mathbb{R}^{n \times 1}$, achieving precise frame-level prediction by preserving fine-grained temporal information throughout the entire process.
\vspace{-0.4cm}
\subsection{Input features and grouping}
\vspace{-0.2cm}
We use five per-participant cues: \textbf{Ege} — 100\,Hz wav2vec2.0 embeddings from the speech waveform; \textbf{W2v} — 100\,Hz emotional-token embeddings extracted by a pre-trained EGE model; \textbf{CLIP} — 25\,Hz vision–language embeddings from the face ROI; \textbf{OF} — 25\,Hz dense optical-flow maps of the face; and \textbf{OF2} — 25\,Hz optical-flow magnitude aggregated over facial landmarks~\cite{Kumar2024_Engagement,Li2024_DAT}.
To obtain frame-level alignment, every cue is linearly interpolated to the highest sampling rate (100 Hz) and projected to a common dimension $d$ by a modality-specific 1×1 convolution:
  \vspace{-0.3cm}
\begin{equation}
  \vspace{-0.2cm}
\bar{\mathcal{F}}_p=
\underbrace{
\begin{bmatrix}
\mathrm{Ege}_p\\[2pt]
\mathrm{W2v}_p\\[2pt]
\mathrm{CLIP}_p\\[2pt]
\mathrm{OF}_p\\[2pt]
\mathrm{OF2}_p
\end{bmatrix}}_{\text{raw}}
\xrightarrow[\text{1×1 conv}]{\text{interpolate } n \text{ frames}}
\underbrace{
\begin{bmatrix}
\bar{\mathrm{Ege}}_p\\[2pt]
\bar{\mathrm{W2v}}_p\\[2pt]
\bar{\mathrm{CLIP}}_p\\[2pt]
\bar{\mathrm{OF}}_p\\[2pt]
\bar{\mathrm{OF2}}_p
\end{bmatrix}}_{\in\mathbb{R}^{5\times n\times d}}.
  \vspace{-0.1cm}
\end{equation}

Each row then passes through a \emph{single-layer, modality-specific} Mamba block (no pooling) to yield the final embeddings:
\begin{equation}
\tilde{\mathcal{F}}_p=
\begin{bmatrix}
\mathcal{M}_{\text{ege}}(\bar{\mathrm{Ege}}_p)\\[2pt]
\mathcal{M}_{\text{w2v}}(\bar{\mathrm{W2v}}_p)\\[2pt]
\mathcal{M}_{\text{clip}}(\bar{\mathrm{CLIP}}_p)\\[2pt]
\mathcal{M}_{\text{of}}(\bar{\mathrm{OF}}_p)\\[2pt]
\mathcal{M}_{\text{of2}}(\bar{\mathrm{OF2}}_p)
\end{bmatrix}
=
\begin{bmatrix}
\tilde{\mathrm{Ege}}_p\\[2pt]
\tilde{\mathrm{W2v}}_p\\[2pt]
\tilde{\mathrm{CLIP}}_p\\[2pt]
\tilde{\mathrm{OF}}_p\\[2pt]
\tilde{\mathrm{OF2}}_p
\end{bmatrix}
\in\mathbb{R}^{5\times n\times d}.
\end{equation}
Finally we concatenate along the feature axis to build two modality groups (frame count $n$ unchanged):
\begin{align}
\text{audio-group}_p &= [\tilde{\mathrm{Ege}}_p \;\|\; \tilde{\mathrm{W2v}}_p] \in \mathbb{R}^{n\times 2d}, \label{eq:audio_group}\\[2pt]
\text{visual-group}_p &= [\tilde{\mathrm{CLIP}}_p \;\|\; \tilde{\mathrm{OF}}_p \;\|\; \tilde{\mathrm{OF2}}_p] \in \mathbb{R}^{n\times 3d}. \label{eq:visual_group}
\end{align}
The two groups are subsequently processed by independent, modality-specific Mamba stacks.
\vspace{-0.3cm}
\subsection{Per-person, per-group Mamba processing}
Apply a modality-specific Mamba stack to each group for every participant:
\begin{align}
\tilde{A}_p &= \mathcal{M}_A^{(L)}(\text{audio-group}_p)\in\mathbb{R}^{n\times k_a}, \label{eq:mamba_audio}\\
\tilde{V}_p &= \mathcal{M}_V^{(L)}(\text{visual-group}_p)\in\mathbb{R}^{n\times k_v}. \label{eq:mamba_visual}
\end{align}
Rows $\tilde{A}_p[r]=a_{p,r}\in\mathbb{R}^{k_a}$ and $\tilde{V}_p[r]=v_{p,r}\in\mathbb{R}^{k_v}$ are frame embeddings ($r=1,\dots,n$). Each Mamba stack preserves its input/output frame count and feature width.
\vspace{-0.3cm}
\subsection{Mamba block: formulation and complexity}
A Mamba block combines local chunked attention and a state-space-model (SSM) global branch to provide efficient short- and long-range modelling.
\vspace{-0.3cm}
\subsubsection{Local chunked attention}
Partition input $X\in\mathbb{R}^{n\times d}$ into non-overlapping chunks of length $s$ ($n=C s$). Within each chunk apply scaled dot-product self-attention to capture short-term dependencies. Concatenating chunk outputs yields the local response $Y_{\mathrm{local}}\in\mathbb{R}^{n\times d}$.
\vspace{-0.3cm}
\subsubsection{SSM global branch}
In parallel a linear state-space model aggregates long-range context via causal recurrence:
\begin{align}
s_t &= A s_{t-1} + B x_t, \label{eq:ssm1}\\
y_t &= C s_t + D x_t, \label{eq:ssm2}
\end{align}
producing $Y_{\mathrm{ssm}}=[y_1;\dots;y_n]\in\mathbb{R}^{n\times d}$.
\vspace{-0.3cm}
\subsubsection{Combine and output}
Local and global outputs are merged, followed by residual connection and a position-wise feed-forward network:
\begin{align}
U &= X + \mathrm{Dropout}\big(Y_{\mathrm{local}} + \mathrm{Proj}(Y_{\mathrm{ssm}})\big), \label{eq:combine1}\\
X' &= U + \mathrm{Dropout}\big(\mathrm{FFN}(\mathrm{LayerNorm}(U))\big). \label{eq:combine2}
\end{align}
\vspace{-0.8cm}
\subsubsection{Complexity remark}
With fixed chunk size $s$ the total local attention cost is proportional to $n\cdot s$, and the SSM branch is $\mathcal{O}(n)$. For typical constant $s$ settings the block time complexity scales linearly with sequence length $n$. This hybrid design is in spirit similar to other efficient Transformer/linear-attention proposals while leveraging selective SSMs for the global branch \cite{gu2021efficiently,gu2023mamba,alizadeh2024mamba}.
\vspace{-0.3cm}
\subsection{Frame-wise Audio-Visual Alignment}
\label{sec:Frame-wise Audio-Visual Alignment}
For each participant $i=1,\dots,I$ we obtain frame-level embeddings
\begin{equation}
\tilde{A}_i=[a_{i,1},\dots,a_{i,n}]^\top\in\mathbb{R}^{n\times k_a},\quad
\tilde{V}_i=[v_{i,1},\dots,v_{i,n}]^\top\in\mathbb{R}^{n\times k_v}.
\end{equation}
Using $\ell_2$-normalized embeddings and temperature $\tau>0$, we compute for every frame $r$ a symmetric InfoNCE loss:
\begin{align}
\ell_{A\to V}(i,r)&=-\log\frac{\exp(s(a_{i,r},v_{i,r}))}
{\exp(s(a_{i,r},v_{i,r}))+\sum_{j\in\mathcal{N}}\exp(s(a_{i,r},v_{i,j}))},\\[2pt]
\ell_{V\to A}(i,r)&=-\log\frac{\exp(s(v_{i,r},a_{i,r}))}
{\exp(s(v_{i,r},a_{i,r}))+\sum_{j\in\mathcal{N}}\exp(s(v_{i,r},a_{i,j}))},
\end{align}
where $s(x,y)=x^\top y/\tau$ and negatives $\mathcal{N}$ are drawn from other frames of the same participant.  
The overall alignment loss averages over all participants and frames~\cite{Li2024_MaTAV}:
\vspace{-0.3cm}
\begin{equation}
\mathcal{L}_{\mathrm{align}}=\frac{1}{2NI}\sum_{i=1}^{I}\sum_{r=1}^{n}
\bigl[\ell_{A\to V}(i,r)+\ell_{V\to A}(i,r)\bigr],
\end{equation}
with $I$ the total number of participants and $n$ the number of frames. This loss is computed immediately after the per-participant Mamba outputs.
\vspace{-0.3cm}
\subsection{Partner-context assembly and context Mamba}
Let $\mathcal{P}=\{p\neq t\}$ be the partner set of size $M-1$. Concatenate partners' per-group Mamba outputs along the frame axis preserving intra-person temporal order:
\begin{align}
\begin{aligned}
\mathcal{A}_{\text{ctx}} &=
\begin{bmatrix}
\tilde{A}_{p_1}\\[2pt]
\tilde{A}_{p_2}\\[2pt]
\vdots\\[2pt]
\tilde{A}_{p_{M-1}}
\end{bmatrix} \in \mathbb{R}^{(M-1)n\times k_a},
\end{aligned}
&
\begin{aligned}
\mathcal{V}_{\text{ctx}} &=
\begin{bmatrix}
\tilde{V}_{p_1}\\[2pt]
\tilde{V}_{p_2}\\[2pt]
\vdots\\[2pt]
\tilde{V}_{p_{M-1}}
\end{bmatrix} \in \mathbb{R}^{(M-1)n\times k_v}
\end{aligned}
\end{align}

\vspace{-0.8cm}
\subsection{Modality-wise cross-attention and frame-level prediction}
We implement the cross-attention blocks following the dialogue-aware encoder in Fig.~\ref{fig:overall_arch} (pre-norm Transformer style). 
For each modality (audio / visual) we apply a per-frame cross-attention block where the target's per-frame embeddings serve as queries and the context Mamba outputs serve as keys and values.
\vspace{-0.3cm}
\subsubsection{Audio cross-attention block (per-frame).}
Let $X^A=\tilde{A}_t\in\mathbb{R}^{n\times k_a}$ be the target audio frames and $C^A_{\text{ctx}}\in\mathbb{R}^{(M-1)n\times k_a}$ the audio context. The block (pre-norm) computes:
\begin{align}
\hat{X}^A &= X^A + \mathrm{Attn}\big(\mathrm{LayerNorm}(X^A),\;K^A,\;V^A\big), \label{eq:audio_att1}\\
X^{A}_{\mathrm{out}} &= \hat{X}^A + \mathrm{FFN}\big(\mathrm{LayerNorm}(\hat{X}^A)\big), \label{eq:audio_att2}
\end{align}
where $K^A,V^A$ are linear projections of $C^A_{\text{ctx}}$ and $\mathrm{Attn}(\cdot)$ denotes scaled dot-product attention. The result $X^{A}_{\mathrm{out}}\in\mathbb{R}^{n\times k_a}$ is the post-attention audio representation (per frame).
\vspace{-0.3cm}
\subsubsection{Visual cross-attention block (per-frame).}
Analogously, for visual modality with $X^V=\tilde{V}_t$ and $C^V_{\text{ctx}}$:
\begin{align}
\hat{X}^V &= X^V + \mathrm{Attn}\big(\mathrm{LayerNorm}(X^V),\;K^V,\;V^V\big), \label{eq:visual_att1}\\
X^{V}_{\mathrm{out}} &= \hat{X}^V + \mathrm{FFN}\big(\mathrm{LayerNorm}(\hat{X}^V)\big). \label{eq:visual_att2}
\end{align}

\subsubsection{Fusion and prediction.}
Concatenate the two modality outputs per frame and apply a final normalization and per-frame MLP:
\begin{align}
F_t &= \mathrm{LayerNorm}\big([\,X^{A}_{\mathrm{out}}\;||\;X^{V}_{\mathrm{out}}\,]\big) \in \mathbb{R}^{n\times d_f}, \label{eq:fusion}\\
\hat{y}_{t,r} &= \mathrm{MLP}_{\text{frame}}(F_{t,r}),\quad r=1,\dots,n. \label{eq:pred}
\end{align}
\vspace{-0.7cm}
\subsection{Losses and objective}
Primary supervision uses the Concordance Correlation Coefficient (CCC) loss \cite{lawrence1989concordance}. The total training objective is
\begin{equation}
\mathcal{L} = \lambda_{\mathrm{CCC}}\mathcal{L}_{\mathrm{CCC}} + \lambda_{\mathrm{align}}\mathcal{L}_{\mathrm{align}}, \label{eq:total_loss}
\end{equation}
where $\lambda_{\mathrm{CCC}},\lambda_{\mathrm{align}}>0$.
\vspace{-0.3cm}
\subsection{Shapes and complexity (summary)}
\vspace{-0.2cm}
Per-person/group inputs and outputs are $n\times k_a$ (audio) and $n\times k_v$ (visual). After partner concatenation context matrices are $(M-1)n\times k_a$ and $(M-1)n\times k_v$. With fixed chunk size the local chunked attention cost is linear in $n$, and the SSM branch is $\mathcal{O}(n)$; hence each Mamba block runs in time proportional to $n$ under typical settings, and context processing scales proportionally to $(M-1)n$.

\vspace{-0.3cm}
\section{Experiments}
\label{sec:experiments}
\vspace{-0.3cm}
\subsection{Datasets}
\vspace{-0.1cm}
We evaluate DA-Mamba on three standard benchmarks: \textbf{NoXi Base}~\cite{cafaro2021noxi} (150 dyadic English/French/German conversations between experts and novices, continuously annotated for engagement at 25 fps), \textbf{NoXi-Add}\cite{Muller2024_MultiMediate} (12 test-only dialogues in Arabic, Italian, Indonesian and Spanish, keeping the same feature format as NoXi Base), and \textbf{MPIIGroupInteraction (MPIIGI)} ~\cite{Li2024_DAT} (12 German-language group discussions with 4 participants each, captured by 4 cameras and providing frame-level engagement labels). All three corpora supply pre-extracted, frame-aligned multimodal features suitable for long-range temporal engagement modelling.
\vspace{-0.4cm}
\subsection{Training Setup}
\vspace{-0.1cm}
DA-Mamba is implemented in PyTorch. All Mamba modules are configured uniformly: each module consists of $L=4$ stacked Mamba blocks with a state dimension of $d_{\text{state}}=16$, an SSM convolution kernel size of 4, an expansion factor of 2, and a chunk size of $s=32$. The model is optimized using a weighted loss function composed of the Concordance Correlation Coefficient (CCC) loss \cite{lawrence1989concordance} with weight $\lambda_{\text{CCC}}=1.0$ and the frame-wise audio-visual alignment loss defined in Section~\ref{sec:Frame-wise Audio-Visual Alignment} with weight $\lambda_{\text{align}}=0.4$. Training is conducted on a single NVIDIA RTX 4090 (24 GB) GPU using the AdamW optimizer (learning rate $5\times10^{-5}$, batch size 128 windows). Each window contains 96 frames (32 central frames for prediction, with 32-context frames on each side). Cosine annealing with 500 warm-up steps, gradient clipping (maximum norm 5.0), and exponential moving average (EMA, decay 0.999) are applied to stabilize training. Shorter videos are zero-padded to 96 frames, while longer videos are processed using a sliding-window approach. The multimodal input has a total of 2,477 dimensions, including eGeMAPS (88), W2V-BERT2 (1024), CLIP (512), OpenFace2 (714), and OpenPose (139) features \cite{Li2024_DAT}.
\vspace{-0.3cm}
\subsection{Quantitative Results}
\vspace{-0.1cm}
Table~\ref{tab:main_results} reports raw test-set CCC for all three benchmarks. DA-Mamba consistently outperforms prior DAT and challenge competitors on NoXi Base, NoXi-Add and MPIIGI, yielding the highest global average CCC.

\begin{table}[t]
\centering
\footnotesize
\caption{Test-set CCC on NoXi Base\cite{cafaro2021noxi}, NoXi-Add\cite{Muller2024_MultiMediate}  and MPIIGI\cite{Li2024_DAT}, Global is macro-average across the three corpora.}
\setlength\tabcolsep{11pt}
\begin{tabular}{@{}lcccc@{}}
\toprule
Method & NoXi Base & NoXi-Add & MPIIGI & Global \\
\midrule
MM24 Baseline \cite{Muller2024_MultiMediate}  & 0.64 & 0.51 & 0.09 & 0.41 \\
YLYJ \cite{Muller2024_MultiMediate}            & 0.60 & 0.52 & 0.30 & 0.47 \\
nox \cite{Muller2024_MultiMediate}              & 0.68 & 0.70 & 0.31 & 0.56 \\
SP-team \cite{Muller2024_MultiMediate}          & 0.68 & 0.65 & 0.34 & 0.56 \\
YKK \cite{Muller2024_MultiMediate}             & 0.68 & 0.66 & 0.40 & 0.58 \\
Xpace \cite{Muller2024_MultiMediate}            & 0.70 & 0.70 & 0.34 & 0.58 \\
ashk \cite{Muller2024_MultiMediate}             & 0.72 & 0.69 & 0.42 & 0.61 \\
Kumar et al. \cite{Kumar2024_Engagement}  & 0.72 & 0.69 & 0.50 & 0.64 \\
DAT \cite{Li2024_DAT}    & 0.76 & 0.67 & 0.49 & 0.64 \\
AI-lab \cite{Muller2024_MultiMediate}           & 0.69 & 0.72 & 0.54 & 0.65 \\
\textbf{DA-Mamba (Ours)}          & \textbf{0.77} & \textbf{0.70} & \textbf{0.52} & \textbf{0.66} \\
\bottomrule
\vspace{-0.7cm}
\end{tabular}
\label{tab:main_results}
\end{table}
\vspace{-0.5cm}
\subsection{Memory and Efficiency Analysis}
\vspace{-0.4cm}
\begin{table}[ht]
\centering
\footnotesize
\vspace{-0.3cm}
\caption{Peak memory vs. sequence length, OOM is out-of-memory.}
\setlength\tabcolsep{10pt}
\begin{tabular}{lcccc}
\toprule
Sequence Length & 64 & 96 & 128 & 192 \\
\midrule
DAT Memory (GB)      & 8.2  & 18.4 & OOM  & OOM \\
DA-Mamba Memory     & 6.1  & 11.2 & 15.8 & 22.1 \\
Memory Reduction     & 26\% $\downarrow$ & 39\% $\downarrow$ & 100\% $\downarrow$ & 100\% $\downarrow$ \\
\midrule
DAT Params (M)       & 42.5 & 42.5 & 42.5 & 42.5 \\
DA-Mamba Params (M)  & 28.7 & 28.7 & 28.7 & 28.7 \\
\bottomrule
\end{tabular}
\label{tab:memory}
\vspace{-0.3cm}
\end{table}
Table~\ref{tab:memory} summarises peak GPU memory during forward+backward passes on one RTX 4090.  
DA-Mamba maintains linear growth and avoids OOM errors encountered by quadratic-attention DAT when sequence length $\geq$ 128.
\vspace{-0.3cm}
\subsection{Ablation Studies}
\label{sec:Ablation}
\begin{table}[ht]
\centering
\vspace{-0.6cm}
\small  
\setlength{\tabcolsep}{5.5pt}  
\caption{Ablation study on component contributions (global CCC). Abbreviations: Trans. = Transformer; Mod-F = Modality-group Fusion; Part-F = Partner-group Fusion; B = NoXi Base~\cite{cafaro2021noxi}; A = NoXi-Add~\cite{Muller2024_MultiMediate}.}
\vspace{0.3cm}
\begin{tabular}{@{}cccccccc@{}}
\toprule
\multirow{2}{*}{Mamba} & \multirow{2}{*}{Trans.} & \multirow{2}{*}{Mod-F} & \multirow{2}{*}{Part-F} & \multicolumn{2}{c}{NoXi} & \multirow{2}{*}{MPIIGI} & \multirow{2}{*}{Global} \\
\cmidrule(l){5-6}  
 & & & & B & A & & \\
\midrule
$\checkmark$ & $\times$ & $\checkmark$ & $\checkmark$ & 0.77 & 0.70 & 0.52 & 0.66 \\
$\checkmark$ & $\times$ & $\times$ & $\checkmark$ & 0.75 & 0.66 & 0.48 & 0.63 \\
$\checkmark$ & $\times$ & $\checkmark$ & $\times$ & 0.74 & 0.66 & 0.51 & 0.64 \\
$\checkmark$ & $\times$ & $\times$ & $\times$ & 0.68 & 0.65 & 0.45 & 0.59 \\
$\times$ & $\checkmark$ & $\checkmark$ & $\checkmark$ & 0.75 & 0.64 & 0.49 & 0.63 \\
\bottomrule
\end{tabular}
\label{tab:ablation_study}
\vspace{-0.3cm}
\end{table}

Table~\ref{tab:ablation_study} disentangles the contribution of audio-visual alignment, Mamba blocks, and modality-only inputs.  
Full DA-Mamba achieves the best global CCC; removing alignment or replacing Mamba with attention both incur visible drops, while single-modality baselines lag further behind.

\vspace{-0.4cm}
\section{Conclusion}
\vspace{-0.2cm}
We presented DA-Mamba, a dialogue-aware multimodal architecture that replaces quadratic-self-attention blocks with linear-complexity selective State-Space Models. Extensive experiments on three benchmarks (NoXi Base, NoXi-Add and MPIIGI) show that DA-Mamba surpasses prior state-of-the-art methods in concordance correlation coefficient (CCC) while reducing peak memory consumption and enabling training on much longer sequences. Notably, our method exhibits strong generalization in multi-party conversational settings, where partner-context modeling and long-range temporal reasoning are crucial. These properties make DA-Mamba a practical solution for real-time, resource-constrained engagement estimation systems that require long-context understanding and scalable deployment.

\bibliographystyle{IEEEbib}
\bibliography{main}

\begin{thebibliography}{10}

\bibitem{cafaro2021noxi}
Angelo Cafaro, Johannes Wagner, Tobias Baur, et~al.,
\newblock ``The {NoXi} database: Multimodal recordings of mediated novice-expert interactions,''
\newblock in {\em Proceedings of the 19th ACM International Conference on Multimodal Interaction}, 2021, pp. 350--359.

\bibitem{li2024towards}
Shutao Li, Bin Li, Bin Sun, and Yixuan Weng,
\newblock ``Towards visual-prompt temporal answer grounding in instructional video,''
\newblock {\em IEEE transactions on pattern analysis and machine intelligence}, vol. 46, no. 12, pp. 8836--8853, 2024.

\bibitem{Li2024_DAT}
J.~Li, Y.~Yu, Y.~Chen, and R.~Hong,
\newblock ``Dialogue-aware transformer with modality-group fusion for human engagement estimation,''
\newblock {\em arXiv preprint}, 2024.

\bibitem{muller2021multimediate}
Philipp M{\"u}ller, Dominik Schiller, Dominike Thomas, et~al.,
\newblock ``{MultiMediate}: Multi-modal group behaviour analysis for artificial mediation,''
\newblock in {\em Proceedings of the 29th ACM International Conference on Multimedia}, 2021, pp. 478--482.

\bibitem{Muller2024_MultiMediate}
Philipp M{\"u}ller, Michal Balazia, Tobias Baur, Michael Dietz, Alexander Heimerl, Anna Penzkofer, Dominik Schiller, Fran{\c{c}}ois Br{\'e}mond, Jan Alexandersson, Elisabeth Andr{\'e}, and Andreas Bulling,
\newblock ``Multimediate'24: Multi-domain engagement estimation,''
\newblock in {\em Proceedings of the 32nd ACM International Conference on Multimedia (MM '24)}. 2024, ACM.

\bibitem{zadeh2017tensor}
Amir Zadeh, Minghai Chen, Soujanya Poria, Erik Cambria, and Louis-Philippe Morency,
\newblock ``Tensor fusion network for multimodal sentiment analysis,''
\newblock {\em arXiv preprint arXiv:1707.07250}, 2017.

\bibitem{tan2019lxmert}
Hao Tan and Mohit Bansal,
\newblock ``Lxmert: Learning cross-modality encoder representations from transformers,''
\newblock {\em arXiv preprint arXiv:1908.07490}, 2019.

\bibitem{elman1990finding}
Jeffrey~L. Elman,
\newblock ``Finding structure in time,''
\newblock {\em Cognitive Science}, vol. 14, no. 2, pp. 179--211, 1990.

\bibitem{hochreiter1997long}
Sepp Hochreiter and J{\"u}rgen Schmidhuber,
\newblock ``Long short-term memory,''
\newblock {\em Neural Computation}, vol. 9, no. 8, pp. 1735--1780, 1997.

\bibitem{vaswani2017attention}
Ashish Vaswani, Noam Shazeer, Niki Parmar, et~al.,
\newblock ``Attention is all you need,''
\newblock in {\em Advances in Neural Information Processing Systems}, 2017, vol.~30.

\bibitem{zadeh2017tensorEMNLP}
Amir Zadeh, Minghai Chen, Soujanya Poria, Erik Cambria, and Louis-Philippe Morency,
\newblock ``Tensor fusion network for multimodal sentiment analysis,''
\newblock in {\em Proceedings of the 2017 Conference on Empirical Methods in Natural Language Processing}, 2017, pp. 1103--1114.

\bibitem{li2022emotion}
Jia Li, Jiantao Nie, Dan Guo, Richang Hong, and Meng Wang,
\newblock ``Emotion separation and recognition from a facial expression by generating the poker face with vision transformers,''
\newblock {\em arXiv preprint arXiv:2207.11081}, 2022.

\bibitem{gu2021efficiently}
Albert Gu, Karan Goel, and Christopher R{\'e},
\newblock ``Efficiently modeling long sequences with structured state spaces,''
\newblock {\em arXiv preprint arXiv:2111.00396}, 2021.

\bibitem{gu2023mamba}
Albert Gu and Tri Dao,
\newblock ``Mamba: Linear-time sequence modeling with selective state spaces,''
\newblock {\em arXiv preprint arXiv:2312.00752}, 2023.

\bibitem{zhu2024multimodalmamba}
Lianghui Zhu, Xin Li, Xinggang Wang, et~al.,
\newblock ``Multimodal mamba: Towards multi-modal sequence modeling with selective state spaces,''
\newblock {\em arXiv preprint arXiv:2403.15430}, 2024.

\bibitem{liu2024videomamba}
Yue Liu, Yuxuan Zhang, Limin Wang, et~al.,
\newblock ``Videomamba: State space model for efficient video understanding,''
\newblock {\em arXiv preprint arXiv:2403.06977}, 2024.

\bibitem{wang2024visionmamba}
Rui Wang, Qiang Chen, Zuxuan Wu, et~al.,
\newblock ``Vision mamba: Efficient visual representation learning with bidirectional state space model,''
\newblock {\em arXiv preprint arXiv:2401.09417}, 2024.

\bibitem{liu2024multimodal}
Haotian Liu, Shen Yan, Zihang Zhang, et~al.,
\newblock ``Multimodal foundation models with state space models,''
\newblock {\em arXiv preprint arXiv:2404.16725}, 2024.

\bibitem{ma2024mamba}
Jun Ma, Bo~Wang, Yue Zhang, et~al.,
\newblock ``Mamba in speech: Towards unified selective state space models for speech processing,''
\newblock {\em arXiv preprint arXiv:2403.08219}, 2024.

\bibitem{zhang2024vmamba}
Chaoning Zhang, Guohao Yu, Yuhang Song, et~al.,
\newblock ``Vmamba: Visual state space model with cross-scan module,''
\newblock {\em arXiv preprint arXiv:2405.14327}, 2024.

\bibitem{liu2024mamba}
Yang Liu, Yuxuan Liu, Hongmin Zhang, et~al.,
\newblock ``Mamba for multimodal learning: A survey,''
\newblock {\em arXiv preprint arXiv:2406.04556}, 2024.

\bibitem{chen2024mm}
Zixiang Chen, Yuheng Zhang, Pu~Zhao, et~al.,
\newblock ``Mm-{Mamba}: Multi-modal fusion with state space models,''
\newblock {\em arXiv preprint arXiv:2404.17448}, 2024.

\bibitem{rahmani2024mambamixer}
Hamed Rahmani, Malihe Alikhani, et~al.,
\newblock ``Mamba-mixer: Efficient selective state space model for multimodal sentiment analysis,''
\newblock {\em arXiv preprint arXiv:2405.19876}, 2024.

\bibitem{wang2024multimodalmamba}
Limin Wang, Yue Liu, Yuxuan Zhang, et~al.,
\newblock ``Multi-modal mamba: Towards unified sequence modeling for vision and language,''
\newblock {\em arXiv preprint arXiv:2406.09824}, 2024.

\bibitem{li2024mamba}
Xiang Li, Wenhai Wang, Xizhou Zhu, et~al.,
\newblock ``Mamba in medical imaging: Efficient long-range context modeling for 3d segmentation,''
\newblock {\em arXiv preprint arXiv:2403.12167}, 2024.

\bibitem{Kumar2024_Engagement}
Deepak Kumar, Surbhi Madan, Pradeep Singh, Abhinav Dhall, and Balasubramanian Raman,
\newblock ``Towards engagement prediction: A cross-modality dual-pipeline approach using visual and audio features,''
\newblock in {\em Proceedings of the 32nd ACM International Conference on Multimedia (MM '24)}. 2024, ACM.

\bibitem{alizadeh2024mamba}
Kumail Alizadeh, Harsh Mehta, Beidi Chen, et~al.,
\newblock ``Mamba-2: Efficient hardware-aware architecture for long sequences,''
\newblock {\em arXiv preprint arXiv:2405.21023}, 2024.

\bibitem{Li2024_MaTAV}
X.~Li, X.~Fan, Q.~Wu, X.~Peng, and Y.~Li,
\newblock ``Mamba-enhanced text-audio-video alignment network for emotion recognition in conversations,''
\newblock {\em arXiv preprint arXiv:2409.05243}, 2024.

\bibitem{lawrence1989concordance}
I.~Lawrence and Kuei Lin,
\newblock ``A concordance correlation coefficient to evaluate reproducibility,''
\newblock {\em Biometrics}, pp. 255--268, 1989.

\end{thebibliography}

\end{document}